\documentclass[10pt,journal,final]{IEEEtran}
\usepackage{mathrsfs}
\usepackage{graphicx}
\usepackage{float}
\usepackage{booktabs}
\usepackage{hyperref}
\usepackage{cite}

\usepackage{enumitem}
\usepackage{color}
\usepackage{psfrag}
\usepackage{subfigure}
\usepackage{amssymb}
\usepackage{amsthm}
\usepackage{setspace}
\usepackage{epsfig}
\usepackage{pifont}
\usepackage{amsmath}
\usepackage{array}
\usepackage{multicol}
\usepackage{multirow}
\usepackage{diagbox}
\usepackage{pifont}
\usepackage{indentfirst}
\usepackage{amsfonts}
\usepackage{algorithm}
\usepackage{algorithmicx}
\usepackage{algpseudocode}
\usepackage{fancyhdr}
\usepackage{amscd}
\usepackage{bm}
\usepackage{fancyhdr}
\usepackage{enumerate}
\usepackage{color}

\newtheorem{proposition}{Proposition}
\newtheorem{definition}{Definition}

\newtheorem{remark}{Remark}
\newcommand{\tabincell}[2]{\begin{tabular}{@{}#1@{}}#2\end{tabular}}

\IEEEoverridecommandlockouts

\begin{document}
	\title{\Large  Learning Power Control for Cellular Systems with Heterogeneous Graph Neural Network}
		
\author{
	\IEEEauthorblockN{Jia Guo and Chenyang Yang}
	
	\IEEEauthorblockA{Beihang University, Beijing, China\\Email: \{guojia, cyyang\}@buaa.edu.cn}\vspace{-2.5em}}
	\maketitle
	
\begin{abstract}
Optimizing power control in multi-cell cellular networks with deep learning enables such a non-convex problem to be implemented in real-time. When channels are time-varying, the deep neural networks (DNNs) need to be re-trained frequently, which calls for low training complexity. To reduce the number of training samples and the size of DNN required to achieve good performance, a promising approach is to embed the DNNs with priori knowledge. Since cellular networks can be modelled as a graph, it is natural to employ graph neural networks (GNNs) for learning, which exhibit permutation invariance (PI) and equivalence (PE) properties. Unlike the homogeneous GNNs that have been used for wireless problems, whose outputs are invariant or equivalent to arbitrary permutations of vertexes, heterogeneous GNNs (HetGNNs), which are more appropriate to model cellular networks, are only invariant or equivalent to some permutations. If the PI or PE properties of the HetGNN do not match the property of the task to be learned, the performance degrades dramatically. In this paper, we show that the power control policy has a combination of different PI and PE properties, and existing HetGNN does not satisfy these properties. We then design a parameter sharing scheme for HetGNN such that the learned relationship satisfies the desired properties. Simulation results show that the sample complexity and the size of designed GNN for learning the optimal power control policy in multi-user multi-cell networks are much lower than the existing DNNs, when achieving the same sum rate loss from the numerically obtained solutions.

\begin{IEEEkeywords}
	\emph{Power control, Graph neural networks, permutation equivalence, permutation invariance, parameter sharing}	
\end{IEEEkeywords}
\end{abstract}

\section{Introduction}
Optimizing power control in multi-cell networks is a well-known non-convex problem. Significant efforts have been devoted to find efficient solutions to this problem. Early works resort to various approximations \cite{MC2017}, which
approximate the problem as a geometric programming problem or convexify the problem by using the monomial approximation for posynomial. In \cite{WMMSE2011Shi}, a weighted sum mean-square error minimization (WMMSE) algorithm was proposed for optimizing coordinated beamforming, which is applicable to the power control problem. However, these numerical or iterative algorithms still incur high complexity if the number of cells is large, which are hard to be implemented for real-time applications.

To avoid solving the optimal solution repeatedly whenever the channels change, an idea of using deep neural networks (DNN) to learn the mapping from channels to the optimal solution has been proposed in \cite{L2O}. While the online computational complexity has been shown reduced remarkably \cite{L2O,PC-CNN}, the off-line complexity of  training is also not negligible. Although this is less of a concern in static scenarios, wireless channels are time-varying. The model parameters and even the size of a DNN have to be updated when channels change. Thus, training DNNs efficiently is critical for wireless applications.


Noticing the fact that generating labels for learning (especially non-convex) optimization problems is time-consuming, unsupervised learning frameworks were proposed for learning to optimize power control in \cite{eisen2018online,PC-ENSEM}, both with fully connected deep neural networks (FC-DNNs). To decrease the required training samples, convolutional neural network (CNN) was applied for power control problem in \cite{PC-CNN}. To allow the well-trained DNN adaptive to the number of users and applicable to large scale systems, graph neural networks (GNNs) were introduced to power control or link scheduling problem in \cite{eisen2019spawc, shen2019graph,lee2020wireless}.
Since cellular networks contain base stations (BSs) and user equipments (UEs) and their relation depends on the channels between BSs and the associated UEs, it is natural to model such a network as a graph and hence apply GNN.

%
%
The architectures of GNNs are embedded with the priori knowledge of graphs, whose vertexes are sets and hence GNNs must be either invariant or equivalent to the permutation of the vertexes \cite{KP2019}.
While GNNs have been demonstrated to achieve good performance in many learning tasks on homogeneous graphs with only one type of vertexes and edges \cite{he2019graph, zhao2020gisnet}, cellular networks are with different types of vertexes (e.g., BSs and UEs) or edges. Such heterogeneous graphs should be learned by heterogeneous GNN (HetGNN). Unlike the homogeneous GNN (HomoGNN) that is invariant or equivalent to arbitrary permutations, HetGNN is only invariant or equivalent to some permutations. If the embedded knowledge in the HetGNN does not match the property of the task to be learned,
the performance will degrade dramatically.

Inspired by the finding that the parameter sharing scheme of a DNN determines the invariance or equivalence relationship it can learn \cite{PEandParamShare2017}, in this paper we design a HetGNN (called PGNN) for learning to optimize power control in multi-cell cellular networks. We first find the permutation invariance
(PI) and permutation equivalence (PE) properties of the relationship between the optimal transmit powers and channels. After modeling the cellular network as a graph, we design a parameter sharing scheme for the HetGNN such that it can learn the desired PI and PE properties. Simulation results show that the sample complexity for training the PGNN is much lower than FC-DNN, vanilla HetGNN, as well as the HomoGNN applied in \cite{lee2020wireless, shen2019graph}, and the model parameters of PGNN are much less than FC-DNN and the HomoGNN.

\emph{Notations:}
$(\cdot)^{\sf T}$ and $(\cdot)^{\sf H}$ denotes transpose and Hermitian transpose, respectively, $|\cdot|$ denotes the magnitude of a complex number, and $\|\cdot\|^2$ denotes two norm.
$\bf \Pi$ or ${\bf \Pi}_m$  denotes a column permutation matrix.

\section{System Model}\label{sec: system model}
Consider a $M$-cell downlink cellular network, where each BS serves multiple UEs with non-real-time service together with other types of services. Each BS is equipped with multiple antennas and each UE is with a  single-antenna. The $m$th BS (denoted as BS$_m$) serves $N_m$ UEs with total allowed power of $P_m^{\max}$, which may be time-varying owing to the dynamic traffic load of different services. If the number of antennas and the coverage are identical among all the BSs, the network is a homogeneous network (HomoNet). Otherwise, it is a heterogeneous network (HetNet), where each marco BS is with more antennas and covers larger cell than each pico BS.

To simplify the analysis, we consider zero-forcing (ZF) beamforming and equal power allocation for the non-real-time UEs, then the data rate of UE$_{n_m}$ over unit bandwidth can be expressed as,
\begin{equation} \label{eq: data rate}
R_{n_m} = \log_2 \left(1 + \frac{|h_{n_m m}|^2p_m/N_m}{\sum_{l=1,l\neq m}^M |h_{n_m l}|^2 p_l/N_l + \sigma_0^2}\right),
\end{equation}
where $p_m$ is the total transmit power to the UEs associated to BS$_m$, $\sigma_0^2$ is the noise power, $h_{j_i m}=\sqrt{\sum_{n_m=1}^{N_m} {\bf g}_{j_i m}^{\sf H} {\bf w}_{n_m} {\bf w}_{n_m}^{\sf H} {\bf g}_{j_i m}}$ is the equivalent channel between UE$_{j_i}$ and BS$_m$ after beamforming, ${\bf g}_{j_i m}$ is the channel vector between BS$_m$ and UE$_{j_i}$, ${\bf w}_{n_m}$ is the beamforming vector of BS$_m$ for UE$_{n_m}$, $\|{\bf w}_{n_m}\|^2=1$ and $\big({\bf g}_{j_m m}\big)^{\sf H}{\bf w}_{n_m}=0, j_m \neq n_m$.

To coordinate inter-cell interference, we optimize the transmit powers at the BSs to maximize the sum-rate of all the non-real-time UEs in the network over unit bandwidth, i.e.,
\begin{subequations}\label{P: max sum-rate}
	\begin{align}
	\max_{p_1,\cdots,p_M} ~~& \textstyle\sum_{m=1}^M\sum_{n_m=1}^{N_m} R_{n_m} \label{P: msr-1} \\
	{\rm s.t.} ~~& 0\leq p_m \leq P_m^{\max}, \label{P: msr-2} \\
	~~& \forall m=1,\cdots,M, n_m=1,\cdots,N_m. \label{P: msr-3}
	\end{align}
\end{subequations}
This problem is non-convex, which can be solved by existing numerical algorithm \cite{MC2017} or iterative algorithm \cite{WMMSE2011Shi}.

To avoid solving the problem repeatedly whenever the channels change, we learn the \emph{optimal power control policy}, i.e., the mapping from the relevant parameters to the optimized transmit powers,
\begin{equation} \label{eq: optimal policy}
{\bf p}^* = F({\bf p}^{\max}, {\bf H}),
\end{equation}
where ${\bf p}^*=[p_1^*,\cdots,p_M^*]^{\sf T}$ is the optimal solution of problem \eqref{P: max sum-rate}, ${\bf p}^{\max}=[P_1^{\max},\cdots,P_M^{\max}]^{\sf T}$, and
\begin{equation} \label{eq: channel matrix}
{\bf H}=
\begin{bmatrix}
{\bf h}_{11} & \cdots & {\bf h}_{1M}\\
\vdots & \ddots & \vdots \\
{\bf h}_{M1} & \cdots & {\bf h}_{MM}
\end{bmatrix},
\end{equation}
with ${\bf h}_{mm'}=[h_{1_m m'},\cdots,h_{n_m m'}]^{\sf T}$. The $m$th column of $\bf H$ contains the channels between all the UEs and BS$_m$, and the $m$th row of $\bf H$ contains the channels between all the BSs and the UEs associated with BS$_m$.

\section{Properties of the Optimal Policy}\label{section IIIproperty}
In this section, we show that the optimal power control policy has a combinational PI and PE properties.
%

We start by providing the definition of several kinds of PI and PE properties to be used in the sequel, including one-dimension (1D)-PI, 1D-PE, two-dimension (2D)-PE and joint-PE.
Consider a function ${\bf Y}=f({\bf X})$, where ${\bf X}=[x_{jk}]$, ${\bf Y}=[y_{jk}]$, $x_{jk}$ and $y_{jk}$ are the elements in the $j$th row of the $k$th column of matrices ${\bf X}$ and $\bf Y$, respectively.

\begin{definition} \label{def: 1D-PI and 1D-PE}
	(1D-PI and 1D-PE) For arbitrary permutation on the rows of $\bf X$, i.e., ${\bf \Pi}^{\sf T}{\bf X}$, if we have ${\bf Y}=f({\bf \Pi}^{\sf T}{\bf X})$, then ${\bf Y}=f({\bf X})$ is 1D-PI to $\bf X$. If we have ${\bf \Pi}^{\sf T}{\bf Y}=f({\bf \Pi}^{\sf T}{\bf X})$, then ${\bf Y}=f({\bf X})$ is 1D-PE to $\bf X$.
\end{definition}

\begin{definition} \label{def: 2D-PE}
	(2D-PE) For arbitrary permutations on the columns and rows of ${\bf X}$, i.e., ${\bf \Pi}^{\sf T}_1{\bf X}{\bf \Pi}_2$, if we have ${\bf \Pi}^{\sf T}_1{\bf Y}{\bf \Pi}_2=f({\bf \Pi}^{\sf T}_1{\bf X}{\bf \Pi}_2)$ or ${\bf \Pi}^{\sf T}_1{\bf Y}=f({\bf \Pi}^{\sf T}_1{\bf X}{\bf \Pi}_2)$, then ${\bf Y}=f({\bf X})$ is 2D-PE to ${\bf X}$.
\end{definition}

\begin{definition} \label{def: joint-PE}
	(Joint-PE) For arbitrary permutation on both the columns and rows of ${\bf X}$, i.e., ${\bf \Pi}^{\sf T}{\bf X}{\bf \Pi}$, if we have ${\bf \Pi}^{\sf T}{\bf Y}=f({\bf \Pi}^{\sf T}{\bf X}{\bf \Pi})$, then ${\bf Y}=f({\bf X})$ is joint-PE to ${\bf X}$.
\end{definition}

\begin{remark}
	1D-PE is a special case of 2D-PE, and joint-PE is a special case of 2D-PE.
\end{remark}

\begin{remark}
	$\bf X$ in these definitions can also be vectors, and $\bf Y$ can also be vectors or scalars, which are special cases of matrices.
\end{remark}

In what follows, we show the PI and PE properties of the optimal power control policy.

\subsubsection{PE property}
The UEs in the network can be divided into multiple subsets, the UEs in each subset associate to the same BS. When the order of BSs changes meanwhile the order of UE subsets change in the same way, only the order of optimal transmit powers changes accordingly while the power control policy (i.e., the mapping, or the multivariate function) remains unchanged. Hence,  the function in \eqref{eq: optimal policy} has PE property. Specifically,
after changing the order of BSs, ${\bf p}^*$, ${\bf p}^{\max}$ and the columns of ${\bf H}$ are permuted to ${\bf \Pi}^{\sf T}{\bf p}^*$,  ${\bf \Pi}^{\sf T}{\bf p}^{\max}$ and ${\bf H}{\bf \Pi}$. After changing the order of UE subsets, the rows of ${\bf H}{\bf \Pi}$ are permuted to ${\bf \Pi}^{\sf T}{\bf H}{\bf \Pi}$. Then, we have
\begin{equation} \label{eq: permuted optimal policy}
{\bf \Pi}^{\sf T}{\bf p}^* = F({\bf \Pi}^{\sf T}{\bf p}^{\max}, {\bf \Pi}^{\sf T}{\bf H}{\bf \Pi}),
\end{equation}
which indicates that the optimal power control policy in \eqref{eq: optimal policy} is 1D-PE to ${\bf p}^{\max}$ and is joint-PE to ${\bf H}$.

\subsubsection{PI property}
For each subset of UEs associated to the same BS, the total transmit power of the BS does not depend on the order of UEs in the subset. Hence, the function in \eqref{eq: optimal policy} has PI property.
The change of the order of the UEs associated to BS$_m$ can be represented by the permutation matrix ${\bf \Pi}_m, m=1,\cdots,M$. After changing the order of UEs, ${\bf h}_{mm'}$ in \eqref{eq: channel matrix} becomes ${\bf \Pi}_m^{\sf T}{\bf h}_{mm'}, m, m'=1,\cdots,M$. Then, we have
\begin{equation} \label{eq: permuted optimal policy 1}
{\bf p}^* = F({\bf p}^{\max}, {\bf \Pi}_1^{\sf T}{\bf h}_{11},\cdots,{\bf \Pi}_M^{\sf T}{\bf h}_{MM}),
\end{equation}
which indicates that the function $F({\bf p}^{\max}, {\bf H})$ is 1D-PI to ${\bf h}_{mm'}$, $m,m'=1,\cdots,M$.

\section{Learning the Optimal Policy with GNN} \label{sec: GNN}
In this section, we first formulate the problem of learning the optimal power control policy as a heterogeneous graph. Then, we introduce vanilla HetGNN and show that its properties do not match the properties of the optimal policy. We proceed to show how to design the HetGNN to satisfy the PI and PE properties of the policy.

\begin{definition} \label{def: HetGraph}
	A graph, denoted as ${\cal G}=({\cal V}, {\cal E})$, consists of a vertex set ${\cal V}$ and an edge set ${\cal E}$. Each vertex and edge belongs to a type, ${\cal A}$ denotes the set of vertex types and ${\cal R}$ denotes the set of edge types.
	When $|{\cal A}|=|{\cal R}|=1$, ${\cal G}$ is a homogeneous graph (HomoG), otherwise, it is a heterogeneous graph (HetG).
\end{definition}\vspace{-1mm}

In a graph ${\cal G}$, each vertex and each edge may be associated with a \emph{feature} and an \emph{action}. Whether or not two vertexes belong to the same type is determined by whether their features are in the same feature space \cite{HetGNN_attention2019Wang}.

In a machine learning task, the learning model (e.g., FC-DNN) is required to learn a function between actions and features.
The problem to learn the optimal policy $F({\bf p}^{\max}, {\bf H})$ can be formulated as a graph, where each BS and each UE is a vertex, respectively, and the channels are edges. The actions and features of the vertexes and the edges are respectively as follows,
\begin{itemize}
	\item[] \hspace{-6mm} \textbf{Actions:}
	\item The action of each BS (say BS$_m$) is its total transmit power $p_m$. The actions of all the BSs can be expressed as a vector ${\bf p}=[p_1,\cdots,p_M]^{\sf T}$.
	\item[] \hspace{-6mm} \textbf{Features:}
	\item The feature of each BS (say BS$_m$) is its available transmit power $P_m^{\max}$. The features of all the BSs is ${\bf p}^{\max}$.
	\item The feature of the edge between BS$_m$ and UE$_n$ (denoted as edge $(m,n)$) is the equivalent channel $h_{mn}$. The features of all the edges can be represented as $\bf H$.
\end{itemize}

\begin{figure}[!htb]
	\centering
	\includegraphics[width=\linewidth]{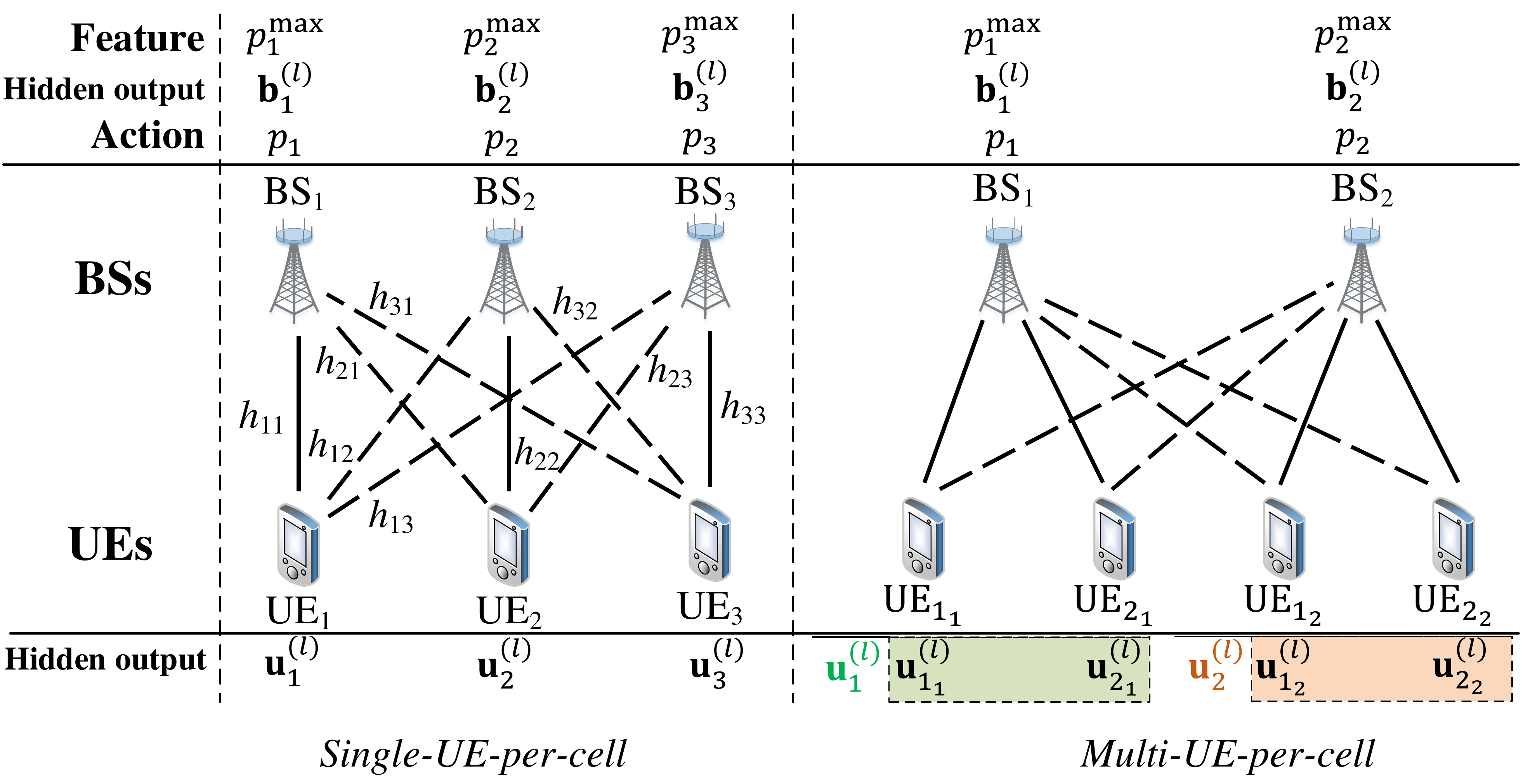}
	\vspace{-2mm}
	\caption{Illustration of WIG and corresponding HetGNN in two cases.}
	\label{fig:fig-systmodel}
	\vspace{-2mm}
\end{figure}

Since BSs and UEs are different types of vertexes, the graph is a HetG, no matter if we consider HetNet or HomoNet. We refer to this graph as \emph{wireless interference graph (WIG)} in the following. In Fig. \ref{fig:fig-systmodel}, we illustrate the features and actions of the WIGs in two cases, as well as the hidden outputs of each UE and BS in the HetGNN (to be explained later).

\vspace{-2mm}\subsection{Heterogeneous Graph Neural Networks}
A GNN contains multiple layers. In each layer, the hidden representations (also called hidden outputs) of all the vertexes are generated. Denote the hidden output in the $l$th layer of the $i$th vertex as ${\bf d}_i^{(l)}$.

HetGNN has been proposed to learn on HetG, where ${\bf d}_i^{(l)}$ is generated with two steps:
\begin{itemize}
	\item[(i)] \textbf{Aggregation}: For all the vertexes neighbored to the $i$th vertex and with the same type (e.g., the $t$th type), their hidden outputs in previous layer (i.e., the $(l-1)$th layer) and the feature of edges connecting the $i$th vertex and the neighbor vertexes are aggregated with an \textbf{aggregator}. The aggregated outputs of the $t$th type of vertexes is,
	\begin{equation}
	{\bf a}_{i,t}^{(l)}
	= {\sf PL}_{j\in{\cal N}_t(i)} \Big(q({\bf d}_j^{(l-1)},{\bf e}_{ij}, {\bf W}_t^{(l)})\Big), t\in{\cal A}, \label{eq: hetgnn aggregator}
	\end{equation}
	where ${\sf PL}(\cdot)$ denotes the pooling function used in the aggregator,
	$q(\cdot,\cdot, {\bf W}_t^{(l)})$ is a parameterized function whose form is determined by the neural network architecture, 	${\bf W}_t^{(l)}$ denotes the model parameters in the $l$th layer for the $t$th type of vertexes that need to be trained, ${\bf e}_{ij}$ is the feature of edge $(i,j)$,
	and ${\cal N}_t(i)$ is the set of vertexes neighbored to the $i$th vertex and with the $t$th type.
	
	\item[(ii)] \textbf{Combination}: After aggregating the information from neighbored vertexes of all types, they are combined with the hidden output of the central vertex (i.e., the $i$th vertex) in the $(l-1)$th layer ${\bf d}_i^{(l-1)}$ to generate  ${\bf d}_i^{(l)}$, by using a \textbf{combiner} as follows,
	\begin{equation}
	{\bf d}_i^{(l)} = {\sf CB}\Big({\bf d}_i^{(l-1)}, \{{\bf a}_{i,t}^{(l)},t\in{\cal A}\}\Big), \label{eq: hetgnn combine}
	\end{equation}
	where ${\sf CB}(\cdot)$ denotes the operation of combination.
\end{itemize}

In \eqref{eq: hetgnn aggregator}, since
the order of neighbors with the same type does not affect each vertex's hidden output, $q(\cdot,\cdot, {\bf W}_t^{(l)})$ is identical for all the vertexes in ${\cal N}_t(i)$, and ${\sf PL}(\cdot)$ is a function satisfying the commutative law, e.g., summation or maximization. We can see from \eqref{eq: hetgnn aggregator} that HetGNN uses different weight matrices when aggregating the information from vertexes of different types, i.e., ${\bf W}_{t}^{(l)}$ differs among types. This is because we need to project different feature spaces to the same space for combining the aggregated information in \eqref{eq: hetgnn combine}.

\vspace{-2mm}\subsection{Properties of the HetGNN for WIG}
For notational simplicity, we first consider the \emph{single-UE-per-cell} case, where ${\bf h}_{ij}=h_{ij}$ is the equivalent channel between UE$_i$ and BS$_j$. Then, the feature of edge $(i,j)$ is ${\bf e}_{ij}={\bf h}_{ij}$. The WIG is a bipartite graph, where each BS is only connected with UEs and each UE is only connected with BSs. Then, \eqref{eq: hetgnn aggregator} and \eqref{eq: hetgnn combine} can be divided into two parts: BSs aggregate information from UEs, and UEs aggregate information from BSs.

For easy understanding, in the sequel we further specify the functions ${\sf CB}(\cdot)$, ${\sf PL}(\cdot)$ and $q(\cdot)$ as the commonly used counterparts in the literature \cite{shen2019graph,lee2020wireless}. To distinguish the hidden outputs between BSs and UEs, we denote ${\bf b}_i^{(l)}$ and ${\bf u}_i^{(l)}$ as the hidden output of the BS$_i$ and UE$_i$, respectively, i.e., ${\bf d}_i^{(l)}$ in \eqref{eq: hetgnn aggregator} and \eqref{eq: hetgnn combine} is replaced by ${\bf b}_i^{(l)}$ or ${\bf u}_i^{(l)}$ when the $i$th vertex is a BS or UE, respectively, as shown in Fig. \ref{fig:fig-systmodel}. Then, for the WIG, \eqref{eq: hetgnn aggregator} and \eqref{eq: hetgnn combine} become
\begin{subequations}\label{eq: hetgnn 1}
	\begin{align}
	\begin{split} \label{eq: hetgnn BS}
	&\textbf{BSs aggregating information from UEs} \\
	&\text{Aggregate:~~} {\bf a}_{i,\text{BS}}^{(l)} = \textstyle\sum_{j=1}^M\left( {\bf V}^{(l)} {\bf u}_j^{(l-1)} + {\bf P}^{(l)} {\bf h}_{ij} \right),\\
	&\text{Combine:~~} {\bf b}_i^{(l)} = \sigma\left( {\bf S}^{(l)} {\bf b}_i^{(l-1)} + {\bf a}_{i, \text{BS}}^{(l)}\right),
	\end{split}\\
	\begin{split} \label{eq: hetgnn UE}
	&\textbf{UEs aggregating information from BSs} \\
	&\text{Aggregate:~~} {\bf a}_{i,\text{UE}}^{(l)} = \textstyle\sum_{j=1}^M\left( {\bf U}^{(l)} {\bf b}_j^{(l-1)} + {\bf Q}^{(l)} {\bf h}_{ji} \right),\\
	&\text{Combine:~~} {\bf u}_i^{(l)} = \sigma\left( {\bf T}^{(l)} {\bf u}_i^{(l-1)} + {\bf a}_{i, \text{UE}}^{(l)}\right),
	\end{split}
	\end{align}
\end{subequations}
where $\sigma(\cdot)$ is an element-wise activation function, e.g., $\texttt{ReLU}$ ($y=\max(x,0)$) or $\texttt{Sigmoid}$ ($y=\frac{1}{1+\exp(-x)}$), ${\bf S}^{(l)}$ and ${\bf T}^{(l)}$ are model parameters in the combination function, ${\bf V}^{(l)}$ and ${\bf U}^{(l)}$ are used to aggregate the information from UEs and BSs, and ${\bf P}^{(l)}$ and ${\bf Q}^{(l)}$ are used to aggregate the information from edges, respectively, which are also the model parameters required to be trained.

Denote ${\bf b}^{(l)}=[{\bf b}_1^{(l){\sf T}},\cdots,{\bf b}_M^{(l){\sf T}}]^{\sf T}$ and
${\bf u}^{(l)}=[{\bf u}_{1}^{(l){\sf T}},\cdots,{\bf u}_{M}^{(l){\sf T}}]^{\sf T}$.
Then, the relationship between hidden outputs of the $l$th and the $(l-1)$th layer can be expressed in matrix form in \eqref{eq: hid out mat 1} (see next page). For notational simplicity, we omit the activation function $\sigma(\cdot)$ and the superscript $(l)$ of weight matrices. Since \eqref{eq: hetgnn BS} and \eqref{eq: hetgnn UE} have the same structure, we only show the relationship where the BSs aggregate information from UEs in \eqref{eq: hid out mat 1}.
In \eqref{eq: hid out mat 1}, $\odot$ is the Hadamard product of two matrices, which makes element-wise product of two matrices and outputs a matrix, ${\bf 1}\triangleq[{\bf I},\cdots,{\bf I}]^{\sf T}$.


From \eqref{eq: hid out mat 1} we can obtain the following Proposition.
\begin{proposition}\label{prop: HetGNN PE PI}
	When learning on WIG with HetGNN, ${\bf b}^{(l)}$ is 1D-PE to ${\bf b}^{(l-1)}$ and 2D-PE to ${\bf H}$.
	\begin{IEEEproof}
		Due to limited space, we omit the proof and only provide the intuition about why the proposition holds. Owing to the parameter sharing in HetGNN, $\bar{\bf S}$ is a block diagonal matrix with identical diagonal blocks, and $\bar{\bf V}, \bar{\bf P}$ are block matrices with all blocks being identical. Hence, for arbitrary permutation of BSs and UEs, when the order of blocks in ${\bf b}^{(l-1)}$, ${\bf u}^{(l-1)}$ and $\bf H$ changes, only the order of blocks in ${\bf b}^{(l)}$ changes but its values remain unchanged.
	\end{IEEEproof}
\end{proposition}

\begin{figure*}[!htb]
	\vspace{-4mm}
	\begin{eqnarray}
	&\overbrace{\begin{bmatrix}
		{\bf b}_1^{(l)}\\
		\vdots\\
		{\bf b}_M^{(l)}
		\end{bmatrix}}^{{\bf b}^{(l)}}
	=\overbrace{\begin{bmatrix}
		{\bf S} & \cdots & {\bf 0} \\
		\vdots & \ddots & \vdots \\
		{\bf 0} & \cdots & {\bf S}
		\end{bmatrix}}^{\bar{\bf S}}
	\overbrace{\begin{bmatrix}
		{\bf b}_1^{(l-1)}\\
		\vdots\\
		{\bf b}_M^{(l-1)}
		\end{bmatrix}}^{{\bf b}^{(l-1)}}
	+
	\overbrace{\begin{bmatrix}
		{\bf V} & \cdots & {\bf V} \\
		\vdots & \ddots & \vdots \\
		{\bf V} & \cdots & {\bf V}
		\end{bmatrix}}^{\bar{\bf V}}
	\overbrace{\begin{bmatrix}
		{\bf u}_1^{(l-1)}\\
		\vdots\\
		{\bf u}_M^{(l-1)}
		\end{bmatrix}}^{{\bf u}^{(l-1)}}
	+
	\overbrace{\begin{bmatrix}
		{\bf P} & \cdots & {\bf P} \\
		\vdots & \ddots & \vdots \\
		{\bf P} & \cdots & {\bf P}
		\end{bmatrix}}^{\bar{\bf P}}
	\odot
	\overbrace{\begin{bmatrix}
		{\bf h}_{11} & \cdots & {\bf h}_{1M} \\
		\vdots & \ddots & \vdots \\
		{\bf h}_{M1} & \cdots & {\bf h}_{M1}
		\end{bmatrix}}^{\bf H}
	\cdot {\bf 1}, \label{eq: hid out mat 1}
	\end{eqnarray}
	\vspace{-3mm}
\end{figure*}

Denote the relationship between the actions ${\bf p}$ and the features ${\bf p}^{\max}$ and $\bf H$ learned by HetGNN as ${\bf p}=G_{\bf W}({\bf p}^{\max},{\bf H})$, where $\bf W$ contains all the model parameters in the HetGNN that need to be trained. Then, the function $G(\cdot)$ has the following PI and PE properties.
\begin{proposition} \label{prop: HetGNN PI PE 2}
	${\bf p}=G_{\bf W}({\bf p}^{\max}, {\bf H})$ is 1D-PE to ${\bf p}^{\max}$, and is 2D-PE to ${\bf H}$.
	\begin{IEEEproof}
		A HetGNN is stacked by multiple hidden layers (e.g., $L$ layers). The first layer is the input layer where the features are inputted, ${\bf b}^{(1)}={\bf p}^{\max}$. The last layer outputs the learned actions, and hence ${\bf b}^{(L)}={\bf p}$ (the corresponding relationship between the actions, features and hidden outputs of the vertexes is shown in Fig. \ref{fig:fig-systmodel}). Since the PI and PE properties in Proposition \ref{prop: HetGNN PE PI} can be preserved by stacking multiple layers  \cite{PEandParamShare2017}, the results in Proposition \ref{prop: HetGNN PI PE 2} hold.
	\end{IEEEproof}
\end{proposition}

\vspace{-2mm}\subsection{Design HetGNN with Desired PE and PI Properties} \label{sec: PGNN}
By comparing the properties of $G_{\bf W}(\cdot)$ in Proposition \ref{prop: HetGNN PI PE 2} and the properties of $F(\cdot)$ given in section \ref{section IIIproperty}, we can see that the PI and PE properties of $G_{\bf W}(\cdot)$ and $F(\cdot)$ does not match. This is because the PI and PE properties of $F(\cdot)$ is determined by the system model (e.g., user association), while the PI and PE properties of $G_{\bf W}(\cdot)$ is determined by the parameter sharing scheme of HetGNN, i.e., the hidden outputs of neighboring vertexes of the same type are aggregated with the same weight matrix (i.e., $\bf U, V, P$ and $\bf Q$) in \eqref{eq: hetgnn 1}.

The training of HetGNN is to search the model parameters $\bf W$ from its hypothesis space such that $G_{\bf W}(\cdot)$ can be as close as $F(\cdot)$. However, when the PI and PE properties of $G_{\bf W}(\cdot)$ and $F(\cdot)$ differ, $F(\cdot)$ may not lie in the hypothesis space of the HetGNN. In this case, $G_{\bf W}(\cdot)$ can never approximate $F(\cdot)$ no matter how well the model parameters ${\bf W}$ are trained.

To resolve this problem, we design a parameter sharing scheme for HetGNN such that $G_{\bf W}({\bf p}^{\max},{\bf H})$ satisfies the PI and PE properties of $F({\bf p}^{\max},{\bf H})$, which is called \textbf{PGNN}.

The relationship between the hidden output of the $i$th vertex in the $l$th layer and the hidden outputs of the $(l-1)$th layer of PGNN is similar to \eqref{eq: hetgnn 1}. Note that the HetGNN has a special architecture where the weight matrices used to aggregate neighbor information are identical (see  $\bar{\bf V}$ and $\bar{\bf P}$ in \eqref{eq: hid out mat 1}), which makes the PE properties of $G_{\bf W}(\cdot)$ does not match the PE properties of $F(\cdot)$. Hence in PGNN, we add matrices ${\bf U, V, P}$ and ${\bf Q}$ with subscripts $ij$ to indicate that the matrices used to aggregate information from neighbored vertexes and edges may differ. For easy exposition, we first consider the case where each BS only serves one user, and then extend to the case where each BS serves multiple users.
\subsubsection{Single-UE-per-cell}
The relationship between the hidden outputs of the $l$th and the $(l-1)$th layers of PGNN can be expressed in matrix form as \eqref{eq: pgnn hid out mat 1}. Again, for notational simplicity, we omit the activation function $\sigma(\cdot)$ and the superscript $(l)$ of weight matrices, and only show the relationship where BSs aggregate information from UEs in \eqref{eq: pgnn hid out mat 1}.

\begin{figure*}[!htb]
	\vspace{-5mm}
	\begin{eqnarray}
	&\begin{bmatrix}
	{\bf b}_1^{(l)}\\
	\vdots\\
	{\bf b}_M^{(l)}
	\end{bmatrix}
	\!\!=\!\!\begin{bmatrix}
	{\bf S} & \cdots & {\bf 0} \\
	\vdots & \ddots & \vdots \\
	{\bf 0} & \cdots & {\bf S}
	\end{bmatrix}
	\begin{bmatrix}
	{\bf b}_1^{(l-1)}\\
	\vdots\\
	{\bf b}_M^{(l-1)}
	\end{bmatrix}
	\!+\!
	\overbrace{\begin{bmatrix}
		{\bf V}_{11} & \cdots & {\bf V}_{1M} \\
		\vdots & \ddots & \vdots \\
		{\bf V}_{M1} & \cdots & {\bf V}_{MM}
		\end{bmatrix}}^{\hat{\bf V}}
	\begin{bmatrix}
	{\bf u}_{1}^{(l-1)}\\
	\vdots\\
	{\bf u}_{M}^{(l-1)}
	\end{bmatrix}
	\!+\!
	\overbrace{\begin{bmatrix}
		{\bf P}_{11} & \cdots & {\bf P}_{1M} \\
		\vdots & \ddots & \vdots \\
		{\bf P}_{M1} & \cdots & {\bf P}_{MM}
		\end{bmatrix}}^{\hat{\bf P}}
	\!\odot\!
	\begin{bmatrix}
	{\bf h}_{11} & \cdots & {\bf h}_{1M} \\
	\vdots & \ddots & \vdots \\
	{\bf h}_{M1} & \cdots & {\bf h}_{MM}
	\end{bmatrix}
	\!\cdot\! {\bf 1}. \label{eq: pgnn hid out mat 1}
	\end{eqnarray}
	\vspace{-2mm}
\end{figure*}

In the following, we design the parameter sharing schemes among ${\bf \hat{U}, \hat{V}, \hat{P}}$ and $\hat{\bf Q}$ such that the relationship between the two layers of PGNN satisfies the properties of $F({\bf p}^{\max},{\bf H})$.
Here, we only discuss the parameter sharing scheme of $\bf \hat{V}$ and $\bf \hat{P}$ in \eqref{eq: pgnn hid out mat 1}, and the parameter sharing of ${\bf \hat{U}}=\{{\bf U}_{ij}\}$ and ${\bf \hat{Q}}=\{{\bf Q}_{ij}\}$ can be designed in the same way.

As shown in section \ref{section IIIproperty}, $F(\cdot)$ is invariant to arbitrary permutation of the elements in ${\bf h}_{ij}$. When each BS only serves one user, however, ${\bf h}_{ij}$ is a scalar with only one element. Therefore, 
it is unnecessary to consider the PI property of $F(\cdot)$ in this case. We only design the parameter sharing scheme to satisfy the PE property of $F(\cdot)$, i.e., $F({\bf p}^{\max},  {\bf H})$ is 1D-PE to ${\bf p}^{\max}$ and joint-PE to ${\bf H}$.

\begin{proposition}\label{prop: PE weight matrix}
	Consider two functions ${\bf y}=f({\bf x})=\sigma({\bf W}{\bf x})$ and ${\bf y}=g({\bf H})=\sigma({\bf W}\odot{\bf H}\cdot {\bf 1})$, where ${\bf y}=[{\bf y}_1^{\sf T},\cdots,{\bf y}_M^{\sf T}]^{\sf T}$, ${\bf x}=[{\bf x}_1^{\sf T},\cdots,{\bf x}_M^{\sf T}]^{\sf T}$ and $\bf H$ is defined in \eqref{eq: channel matrix}. When $\bf W$ contains $M\times M$ sub-matrices with the following structure, ${\bf y}=f({\bf x})$ is 1D-PE to ${\bf x}$, and ${\bf y}=g({\bf H})$ is joint-PE to ${\bf H}$,
	\begin{equation} \label{eq: weight matrix}
	{\bf W}=
	\begin{bmatrix}
	{\bf B} & {\bf C} &\cdots & {\bf C}\\
	{\bf C} & {\bf B} & \cdots & {\bf C}\\
	\vdots & \vdots & \ddots & \vdots \\
	{\bf C} & {\bf C} & \cdots & {\bf B}
	\end{bmatrix},
	\end{equation}
	where ${\bf B}$ and ${\bf C}$ are matrices.
	\begin{IEEEproof}
		Due to limited space, we omit the proof here. The main idea of the proof is that by observing ${\bf \Pi}^{\sf T}{\bf W}{\bf \Pi}={\bf W}$ for arbitrary permutation matrix $\bf \Pi$, we can obtain ${\bf \Pi}^{\sf T}{\bf y}=f({\bf \Pi}^{\sf T}{\bf x})$ and ${\bf \Pi}^{\sf T}{\bf y}=g({\bf \Pi}^{\sf T}{\bf H}{\bf \Pi})$.
	\end{IEEEproof}
\end{proposition}

We can see from Proposition \ref{prop: PE weight matrix} that by letting $\bf \hat V$ and $\bf \hat U$ in \eqref{eq: pgnn hid out mat 1} have the same structure of ${\bf W}$ in \eqref{eq: weight matrix}, ${\bf b}^{(l)}$ is 1D-PE to both ${\bf b}^{(l-1)}$ and ${\bf u}^{(l-1)}$, and joint-PE to $\bf H$, i.e.,
\begin{equation}\label{eq: PE}
{\bf \Pi}^{\sf T}{\bf b}^{(l)} = \sigma({\bf\bar{S}}\cdot {\bf \Pi}^{\sf T}{\bf b}^{(l-1)} + {\bf\hat{V}}\cdot {\bf \Pi}^{\sf T}{\bf u}^{(l-1)} + {\bf\hat{P}}\odot {\bf \Pi}^{\sf T}{\bf H}{\bf \Pi}\cdot {\bf 1}).
\end{equation}
Then, by stacking $L$ layers, we know that ${\bf p}=G_{\bf W}({\bf p}^{\max},{\bf H})$ is 1D-PE to ${\bf p}^{\max}$ and joint-PE to ${\bf H}$, which has the same PE properties as $F({\bf p}^{\max}, {\bf H})$.

\subsubsection{Multi-UE-per-cell} In this case, $\bf \hat V$ and $\bf \hat P$ should still have the structure as in \eqref{eq: weight matrix} such that $G_{\bf W}(\cdot)$ satisfies the PE properties of $F(\cdot)$. Different from the single-UE-per-cell case, we should further design the structure of sub-matrices ${\bf V}_{ij}$ and ${\bf P}_{ij}$ in \eqref{eq: pgnn hid out mat 1} such that $G_{\bf W}(\cdot)$ satisfies the PI properties of $F(\cdot)$, i.e., ${\bf p}^*=F({\bf p}^{\max},{\bf H})$ is 1D-PI to  ${\bf h}_{ij}$.

\begin{proposition}\label{prop: PI weight matrix}
	A function ${\bf y}=f({\bf x})=\sigma({\bf Wx})$ will be 1D-PI to $\bf x$ if $\bf W$ is composed of identical sub-matrices, i.e., with the following structure,
	\begin{equation}\label{eq: PI weight matrix}
	{\bf W}=[{\bf B},{\bf B},\cdots, {\bf B}].
	\end{equation}
	\begin{IEEEproof}
		Since for arbitrary permutation matrix $\bf \Pi$, ${\bf W}{\bf \Pi}^{\sf T}={\bf W}$, we have ${\bf y}=f({\bf x})=\sigma({\bf Wx})=\sigma({\bf W}{\bf \Pi}^{\sf T}{\bf x})=f({\bf \Pi}^{\sf T}{\bf x})$. Hence, ${\bf y}=f({\bf x})$ is 1D-PI to $\bf x$.
	\end{IEEEproof}
\end{proposition}

Similar to \eqref{eq: PE}, we can see from Proposition \ref{prop: PI weight matrix} that by letting the sub-matrices ${\bf V}_{ij}$ and ${\bf P}_{ij}$ be with the same structure as $\bf W$ in \eqref{eq: PI weight matrix} and then substituting them to \eqref{eq: pgnn hid out mat 1},
${\bf b}^{(l)}$ is 1D-PI to  ${\bf h}_{ij}$ and ${\bf u}_{i}^{(l)}, i,j=1,\cdots,M$, Then, by stacking $L$ layers, we know that
${\bf p}=G_{\bf W}({\bf p}^{\max},{\bf H})$ is 1D-PI to ${\bf h}_{ij}$, which is the same as the PI properties of $F({\bf p}^{\max},{\bf H})$.

\begin{figure}[!htb]
	\centering
	\begin{minipage}[t]{0.45\linewidth}	
		\subfigure[PGNN]{
			\includegraphics[width=\textwidth]{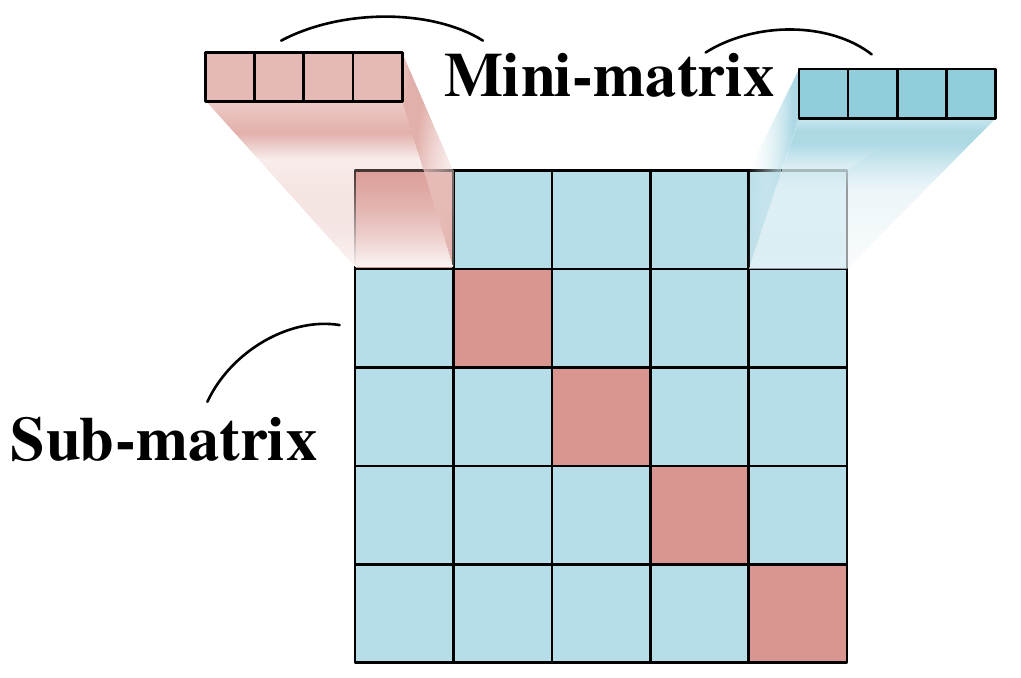}}
	\end{minipage}
	\begin{minipage}[t]{0.45\linewidth}	
		\subfigure[HetGNN]{
			\includegraphics[width=\textwidth]{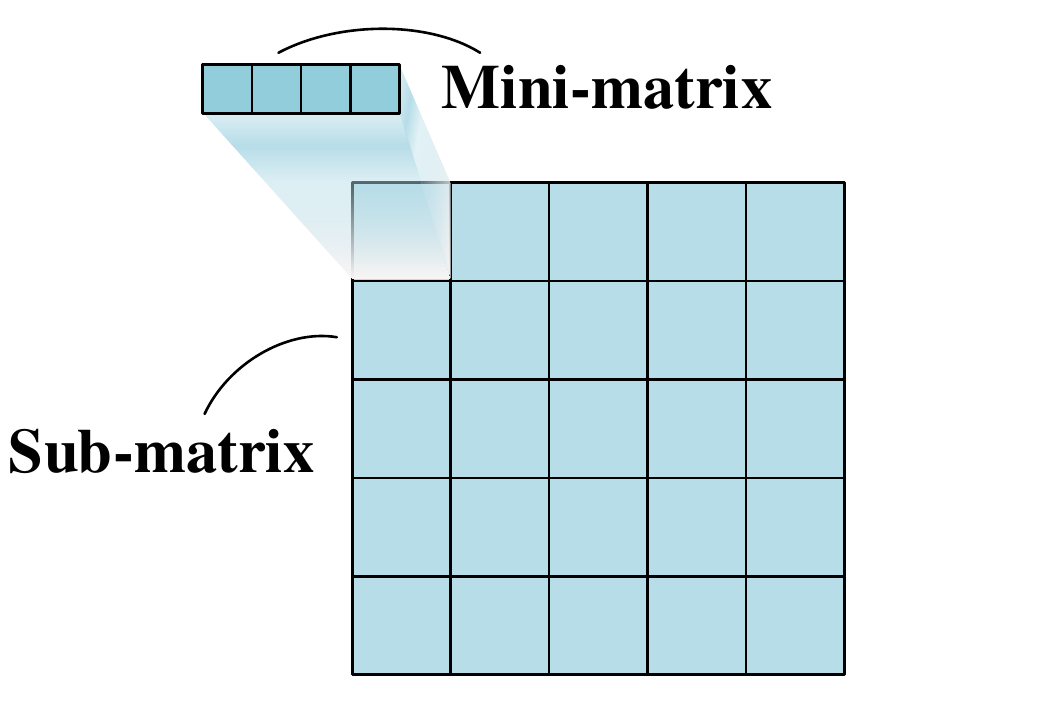}}
	\end{minipage}
	\vspace{-1mm}
	\caption{Parameter sharing scheme of weight matrices, where the same color indicates the same sub-matrix or mini-matrix.}\label{fig:fig-sharemat}
	\vspace{-5mm}
\end{figure}

With the designed parameter sharing scheme, the structure of weight matrices $\bf \hat V, \hat U, \hat P$ and $\bf \hat Q$ in PGNN is shown in Fig. \ref{fig:fig-sharemat} (a).
Each weight matrix contains $M\times M$ sub-matrices, where the sub-matrices are identical on the diagonal and non-diagonal positions, respectively. This parameter sharing scheme aims to guarantee the PE properties of $F(\cdot)$. Each sub-matrix further contains multiple mini-matrices, where all of them are the same. This parameter sharing scheme is to guarantee the PI properties of $F(\cdot)$. For comparison, we also show the structure of weight matrices of the HetGNN in Fig. \ref{fig:fig-sharemat} (b), where all the sub-matrices are identical.

\section{Simulation Results}
Consider a cellular network with $M_{\rm S}$ marco BSs and $M_{\rm P}$ pico BSs. Each marco BS and pico BS is equipped with $N_{\rm S}^{\sf tx}$ and $N_{\rm P}^{\sf tx}$ antennas, and serves $N_{\rm S}$ and $N_{\rm P}$ UEs, respectively, where $N_{\rm S}^{\sf tx}\geq N_{\rm S}$ and $N_{\rm P}^{\sf tx}\geq N_{\rm P}$ such that the multi-user interference can be completely eliminated by ZF beamforming. The channels between UEs and BSs are subject to Rayleigh fading. We compare the performance of the following DNNs.
\begin{itemize}
	\item \textbf{PGNN}: This is the heterogeneous GNN with the parameter sharing scheme we designed in section \ref{sec: PGNN}.
	\item \textbf{HetGNN}: This is the vanilla heterogeneous GNN in section \ref{sec: GNN}-B, where the features of neighboring vertexes of different types, i.e., BSs and UEs, are aggregated with different aggregators.
	\item \textbf{HomoGNN}: This is a homogeneous GNN used in \cite{shen2019graph,lee2020wireless}, which is designed for HomoG. To formulate the considered problem as a HomoG, each BS, all its associated UEs and the channels between them are seen as a vertex, while the interference channels are seen as edges.
	\item \textbf{FC-DNN}: The inputs of this FC-DNN are the features of vertexes and edges, i.e., $\{{\bf p}^{\max}, {\bf H}\}$, and the outputs are the learned actions ${\bf p}^*$.
\end{itemize}

For PGNN, HetGNN and HomoGNN, the inputs and outputs are the features and actions of the corresponding vertexes and edges,  respectively.

The fine-tuned hyper-parameters of the DNNs are shown in Table \ref{table: hyper params}. The number of hidden nodes of PGNN and HetGNN equals to the number of elements in the hidden output vector of each vertex, e.g., ${\bf b}_m^{(l)}$. The activation function of each hidden layer is the commonly used $\texttt{ReLU}$, and the activation function in the output layer is $\texttt{Sigmoid}$ such that the constraint \eqref{P: msr-2} can be satisfied. Each DNN is trained with 1000 epochs.

\begin{table}[htb!]
	\centering
	\vspace{-2mm}
	\caption{Hyper-parameters for the DNNs.}\label{table: hyper params}
	\vspace{-2mm}
	\footnotesize
	\begin{tabular}{c|c|c|c|c}
		\hline\hline
		\multirow{2}{*}{\textbf{Parameters}} & \multicolumn{4}{c}{\textbf{Values}} \\
		\cline{2-5}
		~ & \scriptsize{PGNN} & \scriptsize{HetGNN} & \scriptsize{HomoGNN} & \scriptsize{FC-DNN} \\
		\hline
		\tabincell{c}{\scriptsize Number of hidden layers} & 1  & 1 & 1 & 1 \\
		\hline
		\tabincell{c}{\scriptsize Number of hidden nodes} & 5 & 5 & 10 & 200 \\
		\hline
		\tabincell{c}{\scriptsize Number of model parameters} & 65 & 40 & 420 & 97,808\\
		\hline
		\tabincell{c}{\scriptsize Initial learning  rate} & 0.0005 & 0.0005 & 0.0005 & 0.001\\
		\hline
		\tabincell{c}{\scriptsize Learning algorithm} & \multicolumn{3}{c|}{RMSprop} & Adam \\
		\hline
		\tabincell{c}{\scriptsize Decay rate of learning rate} & 0.9 & 0.9 & 0.9 & --- \\
		\hline
		\tabincell{c}{\scriptsize Back propagation algorithm} & \multicolumn{4}{c}{Iterative batch gradient descent } \\
		\hline\hline
	\end{tabular}
\vspace{-3mm}
\end{table}

Each GNN is trained in a supervised manner to minimize the empirical mean square errors between the outputs of the GNN and the expected outputs over all the training samples. Each sample is composed of an input containing all the features in the graph, i.e., $\{{\bf p}^{\max}, {\bf H}\}$, and an expected output of actions, i.e., optimal transmit power $\bf p^*$ obtained by solving \eqref{P: max sum-rate} with the WMMSE algorithm \cite{WMMSE2011Shi}.

The performance metric is the ratio of the sum-rate achieved by the learned policy to the sum-rate achieved by the WMMSE algorithm, which is called \emph{performance ratio}.

\begin{figure}[!htb]
	\centering
	\includegraphics[width=0.7\linewidth]{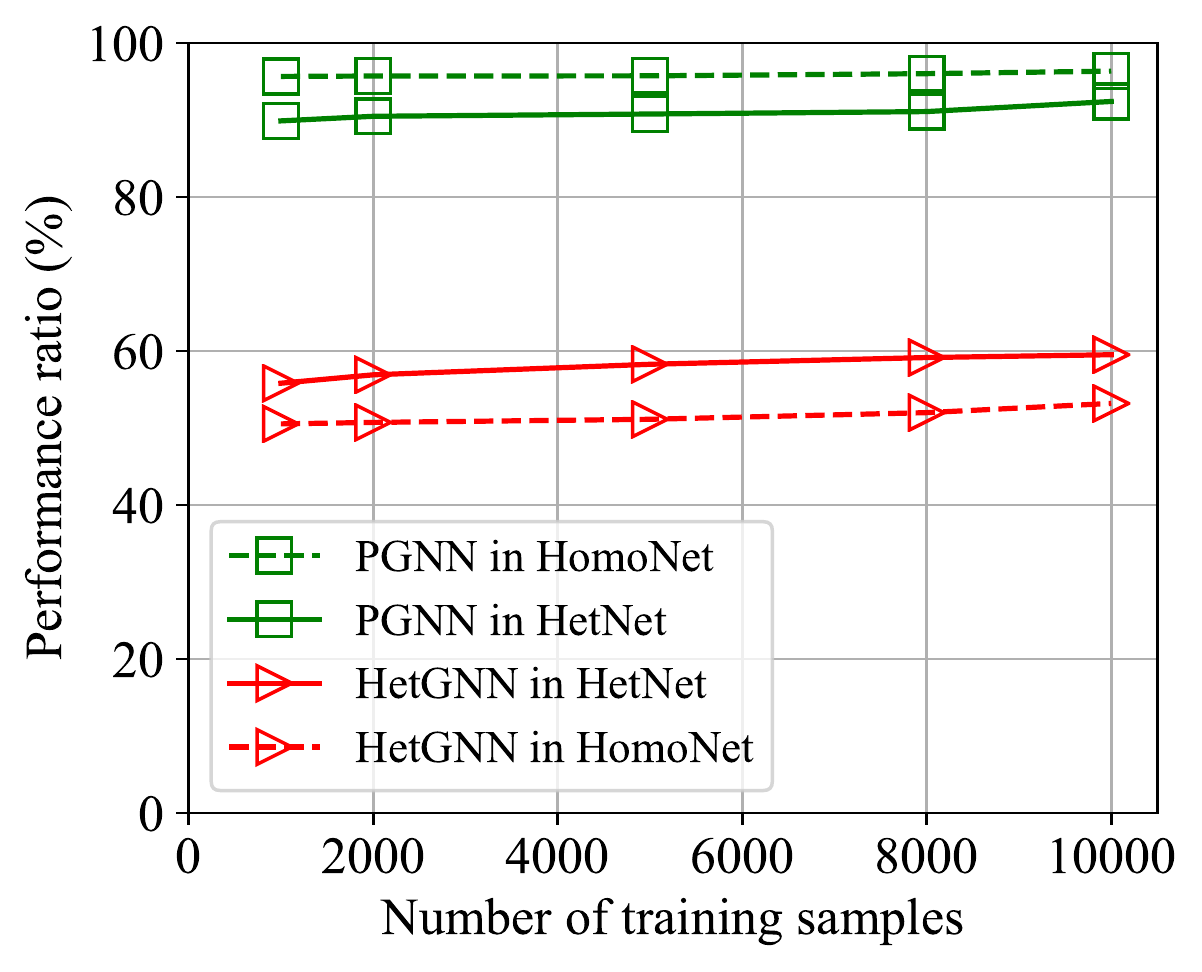}
	\vspace{-1mm}
	\caption{Performance ratio achieved by the two HetGNNs.}
	\label{fig:result}
	\vspace{-1mm}
\end{figure}

In Fig. \ref{fig:result}, we compare the performance ratio of the PGNN and HetGNN for HetNet and HomoNet. In the HetNet, $M_{\rm S}=3, M_{\rm P}=5, N_{\rm S}^{\sf tx}=16, N_{\rm P}^{\sf tx}=8, N_{\rm S}=10, N_{\rm P}=6$, In the HomoNet, there are only marco BSs, and $M_{\rm S}=10, N_{\rm S}^{\sf tx}=16, N_{\rm S}=10$. It can be seen that the performance of PGNN has dramatic gains over HetGNN, because $G_{\bf W}({\bf p}^{\max}, {\bf H})$ of PGNN satisfies the PI and PE properties of $F({\bf p}^{\max}, {\bf H})$, but the function learned by HetGNN does not.

The number of model parameters in a DNN affects the computational complexity in training phase. In Table \ref{table: hyper params}, we compare the number of model parameters  in each DNN, which is obtained in the HetNet scenario. It is shown that much fewer model parameters (and hence the time for training) are required by GNN than FC-DNN to achieve their best performance. PGNN and HetGNN further reduce the model size by more than 80\% with respect to HomoGNN, which indicates the importance of the proper formulation of a problem as a graph. To validate that learning the optimal power control policy allows real-time online implementation, we also compare the running time of the DNN-based solutions in the test phase for 1,000 samples on a computer with Intel CoreTM i9-9940X CPU
(3.30GHz), again  in the HetNet scenario. The running time of the WMMSE algorithm is 14.3 s, and the time of all the DNNs is less than 0.065 s.

In Table \ref{table: sampl-cplxty}, we compare the sample complexities of PGNN, FC-DNN and HomoGNN to achieve the same performance. The sample complexity is defined as the minimal number of training samples required to achieve an expected performance, set as $90\%$ performance ratio for the three DNNs.

It is shown that the sample complexity of PGNN in both cellular networks is much lower than FC-DNN. This is because PGNN incorporates the priori knowledge of the task it intending to learn, hence the hypothesis space for the function to be learned is smaller than FC-DNN.
It is also shown that the sample complexity of PGNN is lower than HomoGNN. This is because HomoGNN models the multi-cell system as a homogeneous graph where each vertex is a combination of a BS, its associated UEs and all the channels between them, which does not fully take the advantage of the PI and PE properties of the considered task. As a result, the feature of each vertex contains the information of all these components, which is with high dimension and hence HomoGNN needs more model parameters as shown in Table \ref{table: hyper params}.

\begin{table}[htb!]
	\centering
	\vspace{-2mm}
	\caption{Sample Complexities of the DNNs}\label{table: sampl-cplxty}
	\vspace{-2mm}
	\footnotesize
	\begin{tabular}{c|c|c|c}
		\hline\hline
		~ & PGNN & HomoGNN & FC-DNN   \\
		\hline
		HetNet &  100 & 300 & 8,000   \\
		\hline
		HomoNet & 50 & 300 & 5,000 \\
		\hline\hline
	\end{tabular}
	\vspace{-5mm}
\end{table}

\section{Conclusions}
In this paper, we learned the optimal power control policy to coordinate inter-cell interference in cellular networks with heterogeneous GNN. We first analyzed the PI and PE properties of the optimal policy, which are used as the priori knowledge to design the GNN. After modeling the multi-cell cellular network as a heterogeneous graph, we designed parameter sharing scheme for heterogeneous GNN such that the learned input-output relationship satisfies the desired PI and PE properties. Simulation results showed that both the sample complexity for training and the model size of the designed GNN are much lower than the existing DNNs.

\bibliography{IEEEabrv,GJ1}
\vspace{-1mm}

\end{document}